\documentclass{article}

\usepackage{PRIMEarxiv}

\usepackage[utf8]{inputenc} 
\usepackage[T1]{fontenc}    
\usepackage{hyperref}       
\usepackage{url}            
\usepackage{booktabs}       
\usepackage{amsfonts}       
\usepackage{nicefrac}       
\usepackage{microtype}      
\usepackage{lipsum}
\usepackage{fancyhdr}       
\usepackage{graphicx}       
\usepackage{xcolor}         
\usepackage{algorithm}
\usepackage{algpseudocode}
\usepackage{amsmath}
\usepackage{cellspace} 
\usepackage{multirow}
\usepackage{subfig}
\usepackage{algorithm}
\usepackage{algpseudocode}
\usepackage{amsmath}
\usepackage{marvosym}

\graphicspath{{media/}}     

\title{Where2Start: Leveraging initial States for Robust and Sample-Efficient Reinforcement Learning
}

\author{
  Pouya Parsa \footnotemark[1]\\
  \texttt{pouya.parsa@aut.ac.ir}
  \and
  Raoof Zare Moayedi \footnotemark[1]\\
  \texttt{raoofmoayedi@aut.ac.ir}
  \and
  Mohammad Bornosi \footnotemark[1]\\
  \texttt{mohho@aut.ac.ir}
  \and
  Mohammad Mahdi Bejani \\ 
  \texttt{mbejani@aut.ac.ir}
}

\newcolumntype{M}[1]{>{\centering\arraybackslash}p{#1}}

\begin{document}
\maketitle
\def\thefootnote{*}\footnotetext[1]{These authors contributed equally to this work}

\begin{abstract}
The reinforcement learning algorithms that focus on how to compute the gradient and choose next actions, are effectively improved the performance of the agents. However, these algorithms are environment-agnostic. This means that the algorithms did not use the knowledge that has been captured by trajectory. This poses that the algorithms should sample many trajectories to train the model. By considering the essence of environment and how much the agent learn from each scenario in that environment, the strategy of the learning procedure can be changed. The strategy retrieves more informative trajectories, so the agent can learn with fewer trajectory sample. We propose Where2Start algorithm that selects the initial state so that the agent has more instability in vicinity of that state. We show that this kind of selection decreases number of trajectories that should be sampled that the agent reach to acceptable reward. Our experiments shows that Where2Start can improve sample efficiency up to 8 times. Also Where2Start can combined with most of state-of-the-art algorithms and improve that robustness and sample efficiency significantly.
\end{abstract}

\section{Introduction}
Despite the remarkable success achieved by deep reinforcement learning (RL), its practical application often encounters significant hurdles, primarily stemming from its vulnerability in real-world settings. The performance of many RL methods tends to falter when confronted with disparities between training and testing scenarios, giving rise to critical safety and security concerns. At the heart of the challenge lies the imperative to develop policies robust enough to withstand disturbances and environmental changes, particularly in addressing the pervasive 'reality gap' problem. This issue encapsulates the formidable task of designing systems capable of effectively navigating the transition from controlled simulated training environments to the intricate and unpredictable conditions of the real world.

One notable limitation in current RL systems is the random selection of initial states for each training episode. This method overlooks the essential characteristics of the environment in the short term, resulting in policies that adeptly handle common scenarios but neglect fundamental yet rare situations and catastrophic phenomena \cite{french1999catastrophic}. Consider a self-driving car navigating through a city devoid of cars and pedestrians. Random placement in the city often allows the agent, maintaining a fixed speed and untouched steering, to achieve a high reward. However, crucial actions such as turning and changing lanes, while less frequent, remain vital. If the agent encounters difficulties in changing lanes but excels at turning, it should be exposed to more lane-changing scenarios than turning scenarios during training. This targeted exposure guides the agent toward mastering more sophisticated maneuvers, minimizing the risk of catastrophic events and enhancing overall learning.

This paper introduces a novel approach, named \textbf{Where2Start}, aimed at training an agent with fewer episodes by strategically selecting initial states where the agent is prone to suboptimal actions. By doing so, the agent is compelled to learn from more challenging scenarios. Additionally, by excluding less informative states, the number of sampled trajectories significantly decreases. Our approach is versatile and can be seamlessly integrated with a wide array of state-of-the-art approaches to achieve superior results in fewer steps. The foundational concept of Where2Start is based on stability. Therefore, we propose a criteria that represents the stability of policy entire states.
\section{Related Work}
\subsection{Robust RL}
The task of formulating robust strategies that can secure high rewards amidst adversarial environmental interferences, a notion commonly known as robust RL, has been examined in scholarly research. This involves creating policies that not only perform well under normal conditions but also maintain their performance when faced with unexpected or adverse changes in the environment
\cite{petrik2019beyond}\cite{russel2020entropic}\cite{derman2021twice}.

Several studies have attempted to leverage adversarial attacks on neural networks to create disturbances in environmental dynamics. For instance, these works utilize the gradient of the critic network to generate disturbances that have meaningful long-term effects. This approach allows for the exploration of how these disturbances impact the performance and robustness of reinforcement learning algorithms in various environments
\cite{pattanaik2017robust}\cite{schott2022improving}\cite{huang2017adversarial}.
However, this line of research does not encompass scenarios or environments where the features of the environment are not directly correlated with observations. This leaves a gap in understanding how reinforcement learning algorithms perform when environmental features and observations are decoupled.

Several studies are exploring ways to improve the robustness of systems by introducing an adversary into the environment, turning the problem into a zero-sum game. This setup is seen as a duel between two players: the protagonist, who is learning to perform a task, and the adversary, who is learning to disrupt the environment to make the protagonist fail. The adversary can cause various disruptions, such as creating disturbances that alter the environment's natural progression and behavior, or manipulating physical obstacles to change the protagonist's navigation routes or strategies. The ultimate goal is to train the protagonist to become more robust against the adversary's disruptions.
\cite{ma2018improved}\cite{pmlr-v37-heinrich15}\cite{heinrich2016deep}\cite{kamalaruban2020robust}\cite{pinto2017robust}.This methodology does present certain challenges. For instance, there isn't a clear correlation between the states of the adversary and the protagonist in these environments. This is akin to the feature engineering of the adversary's states, which can introduce complexity and potential inaccuracies into the system. It's crucial to address these issues to ensure the effectiveness and reliability of the approach.

In our research, we explore a unique scenario where the agent is confronted with complexities in its observations, making it difficult to manage the random noise inherent in these observations. This environment presents unique challenges as the agent is expected to perform efficiently despite the unpredictability and variations introduced by the noise. This situation is particularly relevant as it closely mirrors real-world conditions. It’s not uncommon for an agent’s sensors to experience issues or malfunctions, which can further complicate the agent’s tasks. These complications can arise from various factors such as environmental conditions, hardware limitations, or even software glitches. The agent, therefore, needs to be robust and adaptable, capable of making accurate decisions even when the input data is noisy or incomplete.

\subsection{Sample Efficieny}
The upcoming topic will focus on the issue of how efficiently samples are used in reinforcement learning. The aim in reinforcement learning that is efficient in terms of data usage, especially when dealing with a Markov decision process without any prior knowledge, is to find a policy that achieves a specific goal with the fewest interactions. For a agent, it's a significant challenge to learn an effective policy with the least number of interactions, especially when there's no prior knowledge. Data-efficient methods have been proposed to focus on policy evaluation and enhancement stages. These methods usually use random techniques or fixed strategies to sample initial states. It's important to note that having a more informative initial states can help in learning a more accurate controller, which can lead to fewer interactions. 

In many real world scenarios, each interaction with the environment comes at a cost, and it is desirable for deep reinforcement learning (RL) algorithms to learn with a minimal amount of samples\cite{franccois2018introduction}. Operating real-world physical systems, such as robots, can be expensive, which makes it crucial to learn with the fewest possible number of real-world experiments. This is where the concept of data-efficient reinforcement learning comes into play. It aims to develop a policy that can achieve the desired outcome with minimal interactions, thereby reducing the cost and time associated with numerous trials. The challenge lies in the fact that these systems often lack prior knowledge, making it difficult to establish an effective policy from the outset. However, by utilizing data-efficient methods and having a more informative initial states, we can expedite the learning of a precise controller, ultimately leading to a decrease in the number of required real-world interactions.

The issue of the impact of initial states dates back to the early years of reinforcement learning. In \cite{1617243} a reset distribution has been suggested, which determines the next state based on a specific distribution. Extensive research has been carried out to identify the most critical states for effective training.

Some approaches limit the initial state to a specific set of states. This can be seen, for example, in \cite{salimans2018learning}.
They utilized a collection of demonstration states, initiating each episode by resetting to a state from a demonstration. Interestingly, their agent does not exactly replicate the demonstrated behavior. Instead, it is capable of discovering new and innovative solutions that the human demonstrator might not have considered. This leads to achieving a higher score on Montezuma's Revenge than what was obtained using previously published methods.

Other research has adopted an approach where a memory of the most significant previous states is retained, such as 
\cite{tavakoli2018exploring}\cite{ecoffet2019go}. As an illustration, \cite{ecoffet2019go} presents a technique where the agent's past trajectories are recorded. They then revisit those intersections that seem to hold the potential for revealing new insights, and restart their investigation from those locations.

Moreover, there exists a body of research focused on goal-oriented problems, which employs a "reverse" training approach. This method progressively learns to achieve the goal from a variety of starting points that are incrementally distanced from the goal
\cite{florensa2017reverse}.

Our research stands out from others, especially regarding sample efficiency, due to our unique methodology. We assign a score to each state before initiating each episode, reflecting its level of uncertainty or sensitivity. This strategy is somewhat akin to randomly selecting the initial state, but it yields significantly better convergence than random sampling. We commence exploration from the state with the highest uncertainty or sensitivity, enabling us to develop a policy that concentrates on the most informative or sensitive regions. In addition, we offer a comprehensive framework that accommodates both off-policy and on-policy algorithms. This is a contrast to some previous works that only cater to specific scenarios such as tabular MDPs
\cite{inproceedings}\cite{BARTO199581}\cite{10.1145/1102351.1102423}.
\section{Background}

\subsection{Relative Conditional Number}
The relative condition number is a metric that quantifies the potential variation in the output of a function due to a small perturbation in the input. This measure is particularly useful in environments with noise, as it allows for consistent output generation for an input and its slightly altered, or noisy, version. Essentially, it provides a mechanism to assess the stability of a function's output in the presence of minor fluctuations in the input, making it a valuable tool in maintaining output integrity amidst input uncertainties. For a function $f$ with several variables, we could define the relative condition number as:
\begin{equation}
\frac{\lVert J(X)\rVert.\lVert X\rVert}{\lVert f(X)\rVert}
\end{equation}
where $J(X)$ is the jacobian matrix of partial derivaties of $f$ at $x$.It is important to note that in this study, we have defined the function $f:R^n\rightarrow R$. Consequently, the Jacobian matrix is equivalent to its gradient. Therefore, the relative condition number that we utilize is as follows:
\begin{equation}
\frac{\lVert\nabla f(X)\rVert.\lVert X\rVert}{\lVert f(X)\rVert}
\end{equation}
In the context of our research, we apply the relative condition number in a unique manner. We designate the function 'f' to correspond to the value function, and then calculate the gradient in relation to the parameters of the policy network. This methodology allows us to effectively leverage the relative condition number within our research framework. The expression related to this methodology for each state $s_{t}$ is provided below.

\begin{equation}
    \frac{\lVert\nabla_{\theta} Value_{\pi}(s_{t})\rVert.\lVert s_{t}\rVert}{\lVert Value_{\pi}(s_{t})\rVert}
\end{equation}
Furthermore, In our analysis, we disregard the term $\lVert s_{t}\rVert$, as it does not hold as much significance as the other quantities. The final metric is presented as follows:
\begin{equation}
    \frac{\lVert\nabla_{\theta} Value_{\pi}(s_{t})\rVert}{\lVert Value_{\pi}(s_{t})\rVert}
\end{equation}
\subsection{GP}
Gaussian processes offer a method for directly contemplating the overarching characteristics of functions that could align with our data. They enable us to integrate features such as rapid variation, periodicity, conditional independencies, or translation invariance. The procedure commences with the definition of a prior distribution over potential reasonable functions. This prior is not in search of functions that match the dataset, but rather it aims to specify plausible overarching characteristics of the solutions, like their rate of variation with inputs. Upon data observation, this prior aids in inferring a posterior distribution over functions that could align with the data. Posterior samples can be utilized for making predictions by averaging the values of every potential sample function from the posterior. The posterior mean can be employed for point predictions, and a representation of uncertainty can also be provided to indicate confidence in the predictions. In this study, we employ two metrics to determine the uncertainty or sensitivity of various random states. Utilizing these metrics, we then apply a Gaussian Process (GP) model to fit our data. Essentially, we calculate a score based on these metrics for the entire state space. The state with the highest score is then selected. This approach allows us to systematically evaluate and select the most significant states based on our established criteria.

\subsection{Soft Actor Critic}

SAC incentivizes stochastic exploration through entropy regularization on its policy, this random action noise is insufficient for thoroughly exploring complex environments. Simply increasing entropy leads to undirected state space coverage, wasting samples in already-visited regions. Intelligently searching the state space requires more than injecting random noise into the policy. SAC lacks directed goals for seeking out novel and uncertain states. To enable efficient exploration, the agent needs logic for identifying regions of unpredictability and information gain, setting intrinsic goals for visiting these areas. This targeted search across the state space could uncover new states faster than entropy-driven noise. Overall, SAC's core algorithm does not include mechanisms for focused, information-maximizing exploration. 

\section{Method}
In this section, we detail our methodologies for bolstering the robustness of our model against environmental noise. Despite comprehensive experiments, conventional off-policy reinforcement learning algorithms such as Soft Actor-Critic (SAC) often struggle to achieve robustness against random environmental disturbances. To tackle this, we put forth two distinct strategies. These strategies are specifically designed to enhance the robustness of SAC, thereby making it more resilient to environmental noise. The specifics of these strategies will be elaborated in the following sections.
\subsection{The issue of environmental noise}
To begin with, we try to understand how the model could handle environmental noise. When some perturbation is applied to the agent's observation, the agent should perform as before. Additionally, the agent cannot handle some states in the environment or behave differently in them because it does not learn how to act in them. Seeing a variety of situations is therefore crucial to training an agent effectively and robustly. As a result, the agent is aware of how to behave in different situations, and when there is noise in the observation, which causes a different state for the agent, the agent should know how to react appropriately.  
\subsection{Random selection of initial states}
We propose an initial strategy that involves a more extensive random selection of starting states, marking a departure from traditional methods. While many existing reinforcement learning algorithms also employ random selection, they typically limit their initial states to a confined region within the environment. This restricted scope of random noise may not fully explore the state space, potentially limiting the model's resilience against random disturbances. Our approach, on the other hand, champions a wider selection of initial states with the goal of bolstering the model's robustness. Owing to this comprehensive exploration, our model encounters a variety of situations and is better equipped to handle noisier settings. It's important to note that this broader random selection is not available in existing baseline algorithms.
\subsection{A new metric for measuring the importance of states}

In this section, we focus on a key measure of significance: the Relative condition number. This measure is used to assess the stability of the system. A lower condition number indicates a more stable system, while a higher number suggests potential instability. This metric provides a comprehensive understanding of the system's behavior and performance. To identify the state with the highest score using this method, we initially select various states at random and compute the specified scores for these states. Following this, we implement a Gaussian process on the states to estimate scores across the entire state space. The state yielding the highest score, as determined by the output of the Gaussian process, is then selected. This approach allows us to efficiently explore the state space and identify states of interest based on their scores.\ref{alg:ConditionNumber}
\ref{alg:cap}
\begin{algorithm}[H]
\caption{Calculate Condition Number of Value Function}\label{alg:ConditionNumber}
\begin{algorithmic}
\State \textbf{Input}: Observations state \(S_0\), policy parameters \(\theta\), Q-function parameters \(\phi\), number of sampled actions \(n\)

\State $\mu , \sigma \gets \pi_\theta(S_0)$

\State $\text{sampled actions} \gets \mathcal{N}(\mu, \sigma, \text{n})$

\State $\text{sampled actions probability} \gets \text{calculate probability}(\mu ,\sigma, \text{sampled actions})$

\State $\text{value function} (S_0) \gets \sum_{\text{action in sampled actions}} \text{Q}_\phi(S_0, \text{action}) \cdot \text{sampled actions probability}[\text{action}]$

\State $\text{gradients} \gets  \nabla_\theta \text{value function}(S_0) $

\State \textbf{output:} $\text{condition number} \gets \frac{||\text{gradients}||}{||\text{value function} ||} $

\end{algorithmic}
\end{algorithm}

\begin{algorithm}[H]
\caption{Training on-stability policy gradient}\label{alg:cap}
\begin{algorithmic}
\State \textbf{Input}: state discretization $S_0$, variance threshold $V$, number of epochs $E$
\State $X_\text{train} \gets S_0$
\State $Z_\text{train} \gets \text{normalize} \hspace{1mm} X_\text{train}$
\For{epoch$= E, E-1, \cdots, 1$}
\State $Z_\text{test} \gets \text{shift } Z_\text{train}$
\State $X_\text{train} \gets \text{unnormalize } Z_\text{train}$
\State $Y_\text{train} \gets \text{calculate metric of } \pi(X_\text{train})$
\State $X_\text{test} \gets \text{unnormalize } Z_\text{test}$
\State $Y_\text{test} \gets \text{calculate metric } \pi(X_\text{test})$
\State $\text{model } m \gets \text{fit gaussian process on } Z_\text{train} \text{ and } Y_\text{train}$
\State $\text{variance, mean} \gets \text{inference of } m \text{ on } Z_\text{test}$
\If{$\text{max(variance) } > V$}
\State $i \gets \text{argmax(variance)}$
\State initial\_state $ \gets X_{\text{test}_i} $
\Else
\State $i \gets \text{argmax(mean)}$
\State initial\_state $ \gets X_{\text{test}_i}$
\EndIf
\State obs $ \gets $  reset env with the initial\_state
\While{True}
\State $\text{action} \gets \pi(\text{obs})$
\State $\text{obs, reward, done, }_\text{ } \gets \text{env.step(action)}$
\If{done}
\State break
\EndIf
\EndWhile
\State $\pi \gets \text{update policy's parameters}$
\State $Z_\text{train} \gets Z_\text{test}$
\EndFor

\end{algorithmic}
\end{algorithm}

\section{Experiments}
We designed some experiments to evaluate the performance and measure the impacts of Our proposed method compared to state-of-the-art reinforcement learning algorithms. To demonstrate this enhancement we conducted extensive benchmarking experiments analyzing the performance of SAC
\cite{haarnoja2018soft}
versus On-StabilitySAC using the condition number metric across three distinct environments. The environments examined were the OpenAI benchmarks Pendulum-v1, MountainCarContinuous-v0, and Deepmind's Swimmer-v3. Pendulum-v1 represents a typical environment that is readily solved by common algorithms. MountainCarContinuous-v0 provides a sparse reward landscape, often trapping algorithms in local minima as exemplified by SAC (\ref{fig:sac}). Swimmer-v3 offers a high dimensional observation space inducing substantial computational complexity.\newline
\textbf{Note :} In comparison to the baseline SAC, our method, SAC + RandomSelection, expands the exploration range from the initial state to cover the entire observation space while opting for random selection    
\subsection{Robustness to different noise regimes }
\begin{figure}[H]
\centering
\subfloat[]{\includegraphics[width=0.5\textwidth]{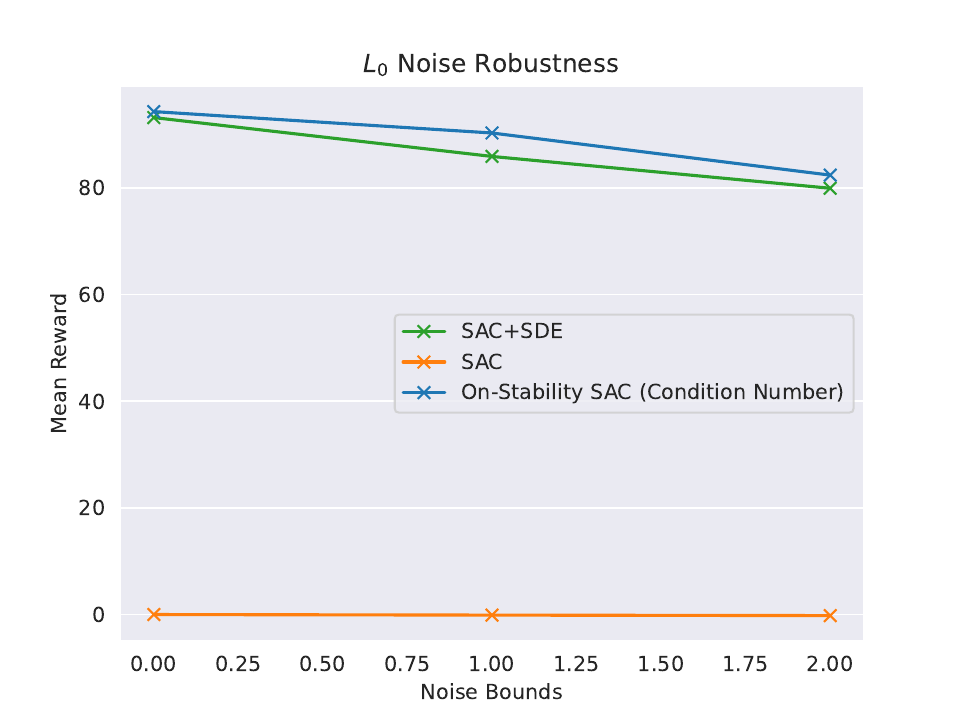}}\hfill
\subfloat[]{\includegraphics[width=0.5\textwidth]{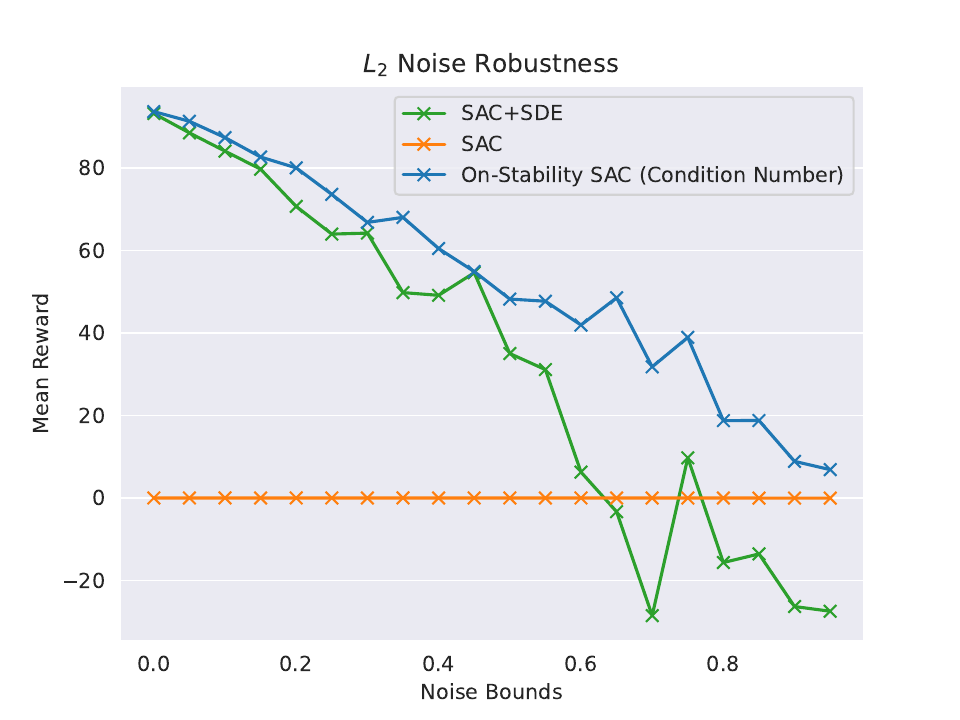}}\
\subfloat[]{\includegraphics[width=0.5\textwidth]{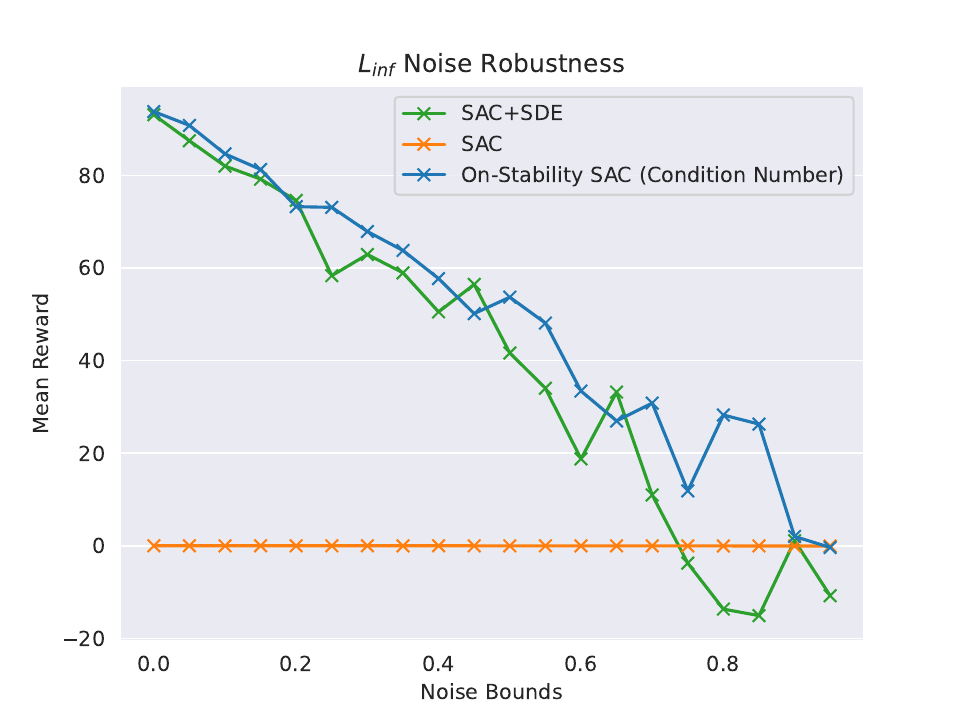}}\hfill
\subfloat[]{\includegraphics[width=0.5\textwidth]{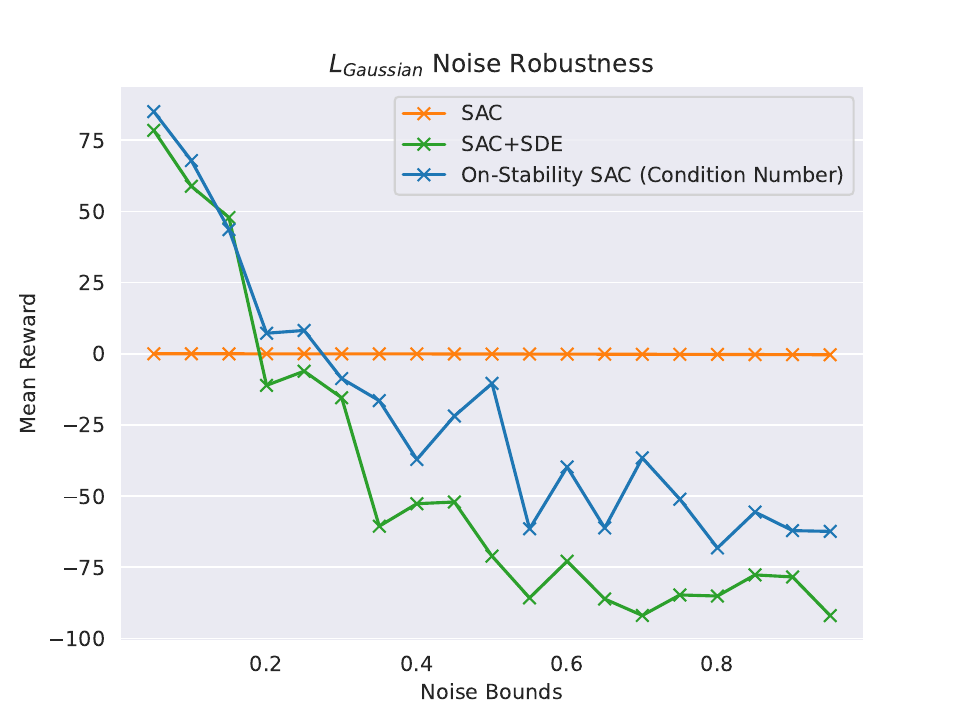}}

\caption{Our experiments underscored the challenge of noise robustness in MountainCarContinuous, a sparse reward environment. Despite the low dimensional observation space, the SAC algorithm failed to identify the global minimum in MountainCarContinuous. This inability to converge highlights the difficulty reinforcement learning algorithms face in overcoming local minima and sparse rewards. Our model demonstrates superior noise resilience in this setting, reaching the global minimum under L0, L2, Linfinity, and Gaussian noise regimes.}
\label{fig:sac}
\end{figure}

Performance of SAC versus On-StabilitySAC using the condition number metric across the mentioned environments under four different noise regimes.  The noise types tested were $L_{0}$, $L_{2}$, $L_{\text{infinity}}$, and Gaussian noise, with the first three representing Lp noise models. These experiments aimed to demonstrate our model's robustness across diverse noise types. Overall, our contributions illustrate enhanced stability and noise resistance compared to established RL techniques on representative benchmark environments and noise models.\ref{fig:sac}

\subsection{Effect of exploration metric}
\begin{figure}[H]
\centering

\subfloat[Pendulum]{\includegraphics[width=0.33\textwidth]{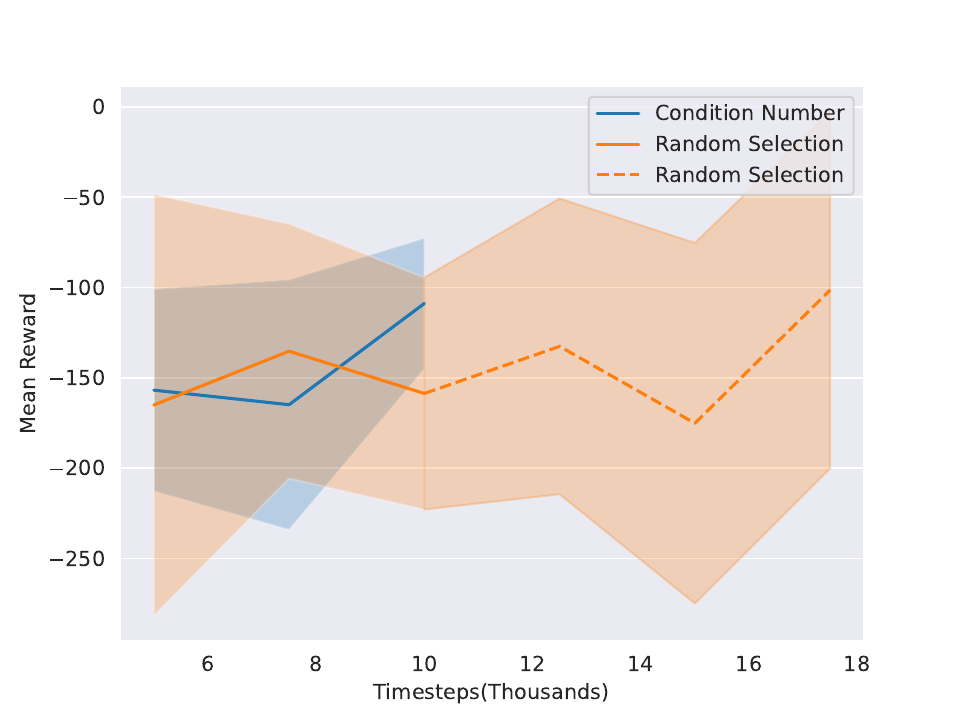}}\hfill
\subfloat[MountainCar]{\includegraphics[width=0.3\textwidth]{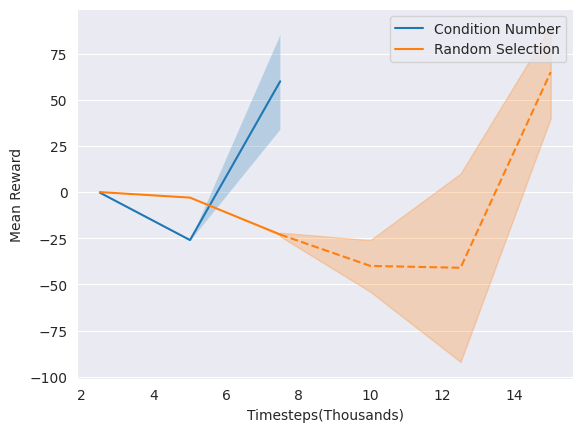}}\hfill
\subfloat[Swimmer]{\includegraphics[width=0.33\textwidth]{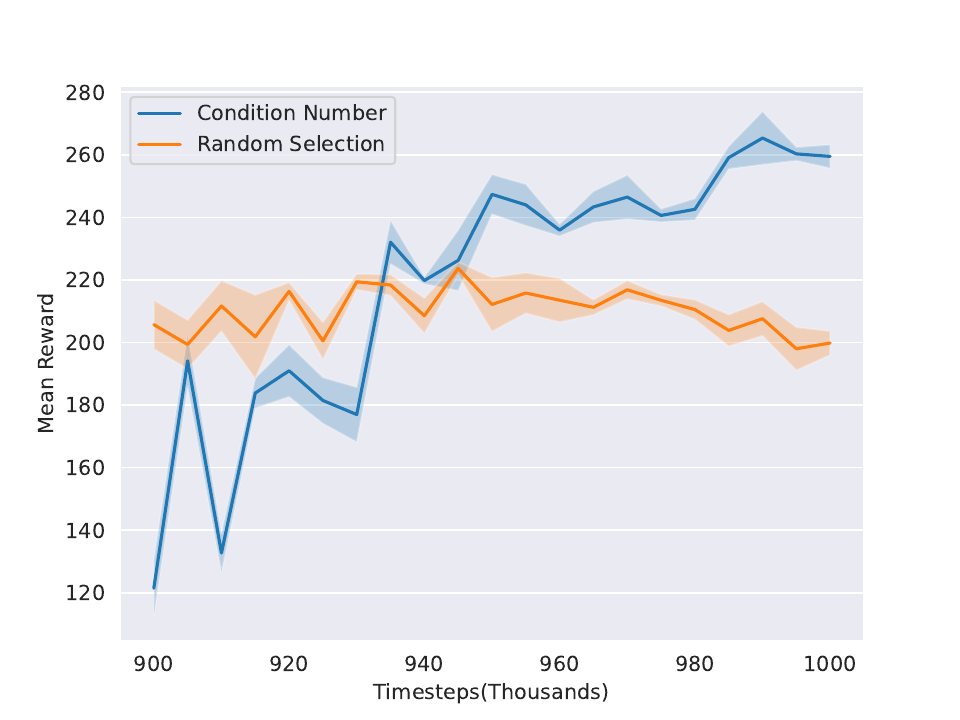}}

\caption{We compared the convergence rates across metrics in Pendulum-v1, MountainCarContinuous-v0, and Swimmer-v3 benchmark environments. The x-axis indicates training time in thousands of timesteps, while the y-axis shows the mean cumulative reward over 100 evaluation episodes. The condition number metric achieves superior cumulative rewards given equal interactions with the environment. Dashed lines demonstrate the substantial additional training time required for alternative metrics to reach parity with the condition number's performance.}
\label{fig: exploration metric}
\end{figure}

Our results indicate that the choice of exploration metric significantly impacts measured performance. We compared our model using two approaches for selecting initial states: random selection versus selection based on the condition number metric. This benchmarking demonstrated substantially different outcomes depending on the exploration metric used. When initial states were chosen randomly, our model showed modest improvements compared to baseline algorithms. However, conditioning initial states on the condition number revealed dramatic enhancements in stability and noise resilience.\ref{fig: exploration metric}

\subsection{Scalability and Sample efficiency }
\begin{figure}[H]
\centering
\subfloat[Swimmer]{\includegraphics[width=0.32\textwidth]{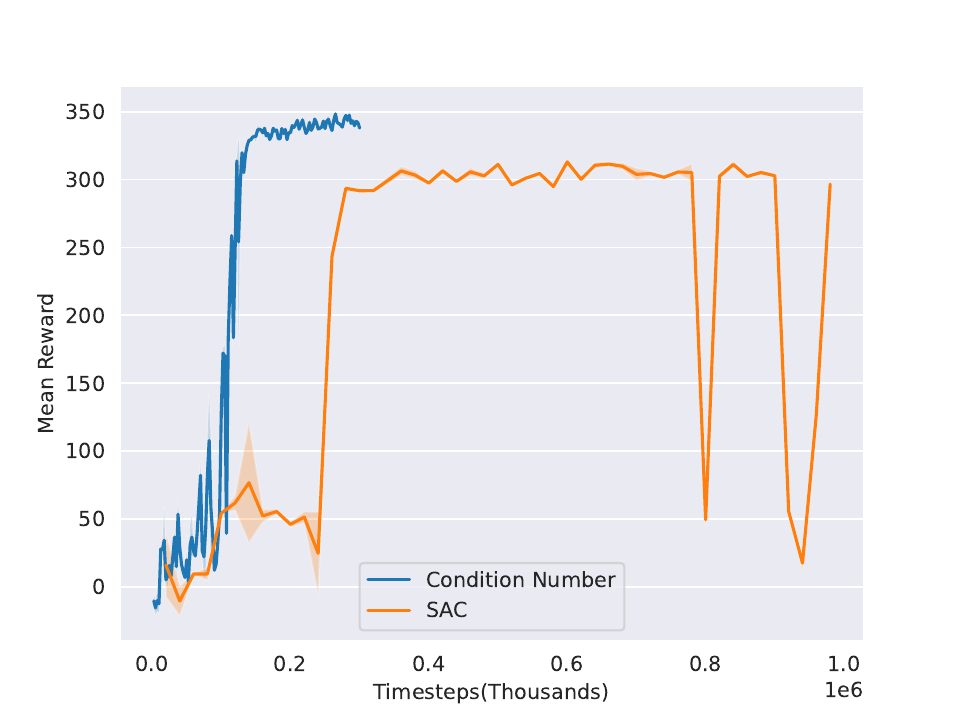}}\hfill
\subfloat[MountainCarContinuous]{\includegraphics[width=0.32\textwidth]{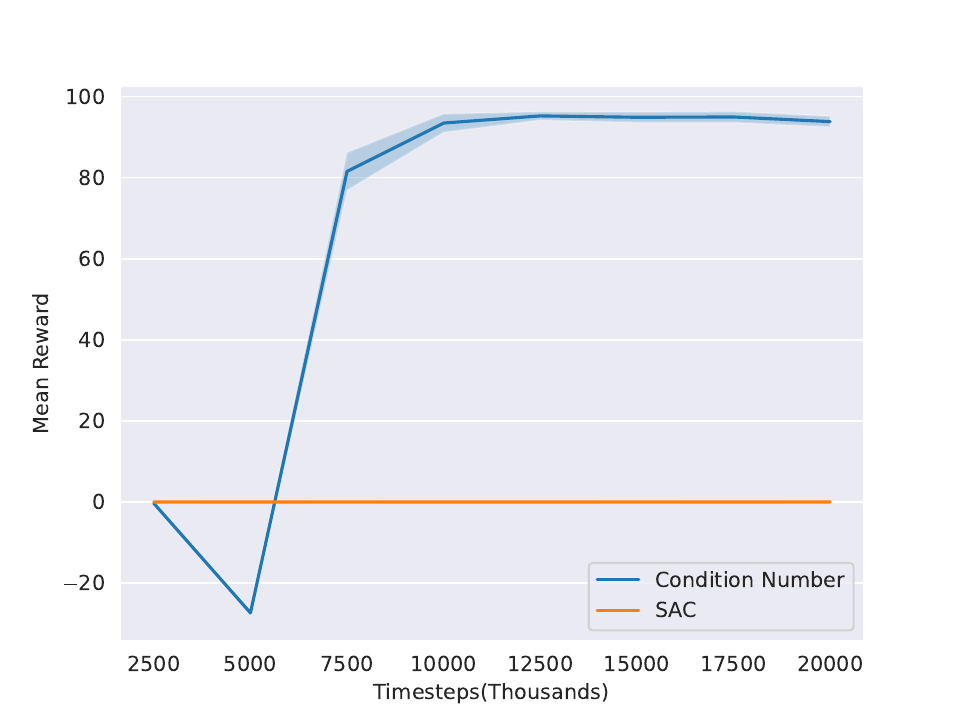}}
\subfloat[Pendulum]{\includegraphics[width=0.32\textwidth]{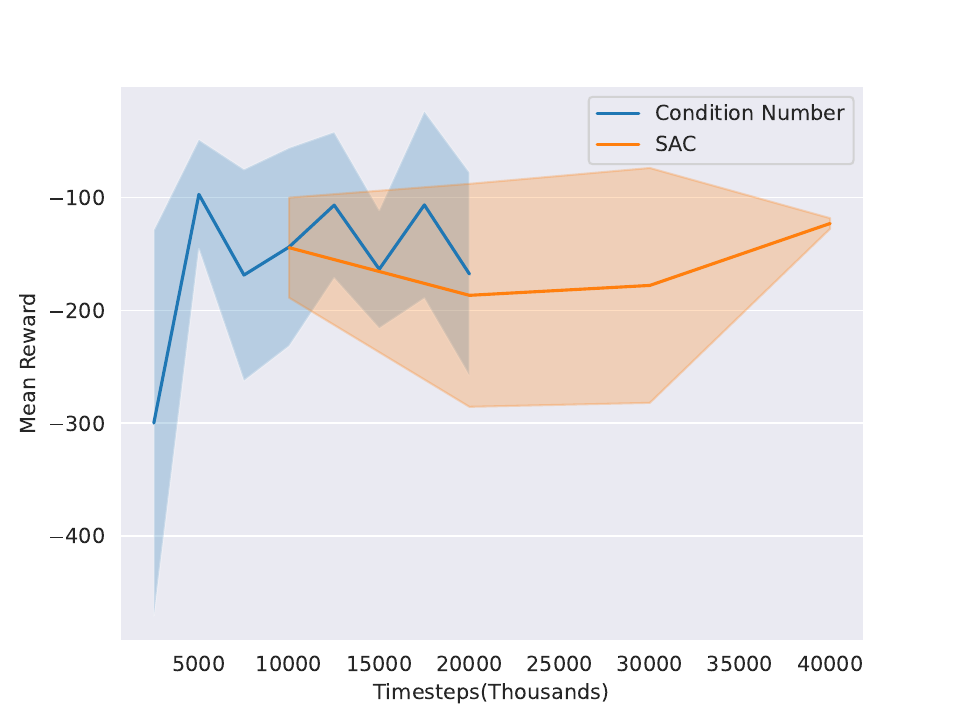}}
\caption{ On-Stability SAC achieves substantially higher cumulative rewards over the course of training compared to SAC. Additionally, On-Stability SAC reaches its maximum performance in approximately 150,000 timesteps, while SAC requires over 300,000 interactions to converge. These results validate our approach's ability to attain superior asymptotic performance with fewer environmental interactions on complex tasks. The condition number metric enables more efficient learning, allowing our algorithm to surpass baseline SAC's performance with nearly 50\% less experience.}
\label{fig:exp_vs_plan}
\end{figure}

Duo to evaluate our model's data efficiency we take advantage of the training curves of our proposed method, on-stability SAC utilizing the condition number metric compared to the baseline SAC conditioning initial state on  Random selection on both the Swimmer and MountainCarContinuous environments.\ref{fig:exp_vs_plan} 

Also by having the advantage of sample efficiency search, we have to compare the order of computation time per one episode of models, to ensure how scalable methods are versus complex environments.\ref{tab: computation time table} 

\begin{table}[H]
  \centering
  \begin{tabular}{|c|c|c|c|c|}
    \cline{2-5}
    \multicolumn{1}{c|}{} &\multicolumn{2}{c|}{Computation Time} & \multicolumn{2}{c|}{Sample Efficiency} \\
    \hline
    Environments & Random Selection & Condition Number & Random Selection & Condition Number  \\
    \hline
    Pendulum-v1 & X & 15X & A & 2A \\
    MountainCarContinuous-v0 & Y & 4.85Y & - & - \\
    Swimmer-v3 & Z & 95Z & B & 8B \\ 
    \hline
  \end{tabular}
  \vspace{8pt}
  \caption{On-Stability SAC + ConditionNumber, while exhibiting sample efficiency in search environments, is associated with a high computation time due to the complexity of environments. This complexity could potentially overshadow data efficiency in search, making it important to consider in particularly complex environments.}
  \label{tab: computation time table}
\end{table}

A comprehensive examination of environmental complexity, spanning from Pendulum-v1 to Swimmer-v3, reveals a heightened density in the observation space due to an increased number of dimensions. Consequently, the exploration of numerous states becomes essential to identify an optimal initial state. (It is noteworthy that the quantity of states considered for metric calculation exhibits an exponential relationship with the observation space's dimension.) As a result, the computational cost associated with our metric calculation is conditioned upon the extensive application of the backward() function on the value function of each state. This, in turn, leads to significant time-consuming training. Nonetheless, there is a conceptual approach wherein leveraging the backward function throughout the training process by baseline algorithms enables the calculation of gradients for the value function, allowing for their efficient \textbf{reuse}.

Now, by incorporating sample efficiency and scalability through the \textbf{reuse} concept, we can assert that our method is poised to be applied effectively in complex environments.

\section{Conclusion}
 In this study, we introduced the innovative approach \textbf{Where2Start} to tackle pivotal challenges in deep reinforcement learning (RL), with a particular focus on mitigating vulnerabilities during the transition from simulated training environments to real-world applications. The persistent 'reality gap' dilemma highlights the imperative for RL systems to dynamically adapt to the intricacies and unpredictabilities inherent in diverse scenarios.

Where2Start takes a strategic approach to selecting initial states for training episodes based on the agent's suboptimal actions, providing a promising avenue for bolstering RL robustness. The method systematically exposes the agent to challenging scenarios influenced by conditional numbers, fostering a more comprehensive understanding of its environment.

Moreover, Where2Start contributes to efficiency gains by significantly reducing the number of sampled trajectories. This efficiency, coupled with the adaptability of our approach to seamlessly integrate with various state-of-the-art methods, positions Where2Start as a valuable tool for achieving superior RL results in a condensed number of training steps. Notably, our experiments reveal that Where2Start can converge to a better sub-optimal agent up to 8 times faster than common approaches. However, it is essential to acknowledge that the computation cost of the conditional number at the start of each episode is up to 95 times higher than common approaches, a critical consideration for future research endeavors.

The emphasis on stability within our approach establishes a solid foundation for ongoing advancements in RL research and applications. As we navigate the evolving landscape of artificial intelligence, Where2Start signifies a significant stride toward more reliable, safer, and efficient RL systems, setting the stage for broader and more impactful real-world implementations.

\bibliographystyle{unsrt}  
\bibliography{references}

\end{document}